\RequirePackage{fix-cm}
\documentclass[smallextended]{svjour3}       
\smartqed  
\usepackage{graphicx}
\usepackage{amssymb,graphics,subfigure}
\usepackage{amsmath}
\usepackage{amsfonts}  
\usepackage{floatflt}
\usepackage{indentfirst}
\usepackage{latexsym,bm,amssymb}
\usepackage{rotating}
\usepackage{multicol,multirow,booktabs}
\usepackage{algorithm}
\usepackage{caption}  
\usepackage{url}
\usepackage{epstopdf}
\begin{document}


\title{Simultaneous merging multiple grid maps using the robust motion averaging}

\author{Zutao Jiang$^1$ \and
        Jihua Zhu $^{1*}$  \and
        Yaochen Li$^1$  \and
        Zhongyu Li$^2$  \and
        Huimin Lu$^3$
}


\institute{Corresponding Author: Jihua Zhu \at
              \email{zhu@xjtu.edu.cn}           
           \and
           1. School of Software Engineering, Xi'an Jiaotong Universiy, P. R. China\at
           \and
           2. Department of Computer Science, University of North Carolina at Charlotte, USA \at
           \and
           3. Kyushu Institute of Technology, Japan \at
}

\date{Received: date / Accepted: date}

\maketitle

\begin{abstract}
Mapping in the GPS-denied environment is an important and challenging task in the field of robotics. In the large environment, mapping can be significantly accelerated by multiple robots exploring different parts of the environment. Accordingly, a key problem is how to integrate these local maps built by different robots into a single global map. In this paper, we propose an approach for simultaneous merging of multiple grid maps by the robust motion averaging. The main idea of this approach is to recover all global motions for map merging from a set of relative motions. Therefore, it firstly adopts the pair-wise map merging method to estimate relative motions for grid map pairs. To obtain as many reliable relative motions as possible, a graph-based sampling scheme is utilized to efficiently remove unreliable relative motions obtained from the pair-wise map merging. Subsequently, the accurate global motions can be recovered from the set of reliable relative motions by the motion averaging. Experimental results carried on real robot data sets demonstrate that proposed approach can achieve simultaneous merging of multiple grid maps with good performances.
\end{abstract}

\section{Introduction}
\label{intro}

Mapping is one of the most fundamental and difficult issues in robotics, and has attracted more and more attention since the seminal work presented in \cite{Smith87}. In the past few decades, many effective approaches \cite{Thrun05} have been proposed to build several kinds of environment maps, such as grid map \cite{Giorg07}, feature map \cite{John11}, topological map \cite{Lui12}, and hybrid map \cite{Bibby10}, etc. As a kind of probabilistic map, the occupancy grid map is not required to extract any special features from environments, so it can easily model arbitrary types of environments. Therefore, the grid map is one of the most popular map representations in robot mapping. However, most of robot mapping approaches can only build single map for medium scale environments. For the large scale environment, multi-robots should cooperatively explore various parts of the same environment so as to build grid map with good efficiency and accuracy. The key problem is how to integrate these local grid maps built by multiple robots into a single global map.

To merge a pair of grid maps, Carpin $et$ $al.$ viewed it as the optimization problem \cite{Carpin05}, where
the optimal transformation should be searched to align two grid maps to be merged. Subsequently,
two stochastic search approaches were proposed to solve this optimization problem \cite{Carpin05,Carpin06}. Similarly, Li $et$ $al.$ proposed an grid map merging approach based on the genetic algorithm \cite{Li14}. Although these approaches may obtain the optimal rigid transformation, they are all time-consuming due to the nature of exhaustive search. Different from these passive merging approaches, some researchers proposed
when two robots meet randomly or search each other out during the mapping, they can perform the map merging by determining their relative pose \cite{Howard06,Fox06}. What's more, Carpin $et$ $al.$ then proposed map merging approach based on the Hough transform \cite{Carpin08}, which can merge grid maps containing the line features. Although this approach can efficiently merge grid map without any line feature extraction, its accuracy should be further improved due to the nature of discretization error in the Hough transform. Besides, it is required that the potentially being merged grid maps should contain a significant overlapping percentage. To address the accuracy issue, Zhu $et$ $al.$\cite{Zhu13} viewed the grid map merging as the point set registration problem and accomplished it by the trimmed iterative closest point (TrICP) \cite{Chet05,Phils07}, where the initial parameters are provided by the map merging approach based on the Hough transform. Meanwhile, Blanco $et$ $al.$ proposed a multi-hypothesis method to provide the initial parameters for point set registration algorithm so as to merge grid maps \cite{Blanco13}. By the confirmation of merging hypotheses, it can obtain the robust merging result. To address the robustness issue, Saeedi $et$ $al.$ proposed the improved grid map merging approach based on the Hough transform, which can merge grid map pair even with low overlapping percentage \cite{Saee14}. To merge grid maps with different resolutions, Ma $et$ $al.$ put forward an image registration based approach \cite{Ma16}, which can determine whether one of the two maps should be minified or magnified in order to be merged with the other. It seems that many proposed approaches can merge grid map pair with good accuracy and efficiency, but few merging approaches can really accomplish simultaneous merging multiple grid maps.

Suppose there is a set of unordered grid maps, which are built by multiple robots exploring different parts of the same large environment. These grid maps are non-overlapping or partially overlapping with each other. Given the reference map, the goal of multiple grid map merging is to integrate these local grid maps into a global map by calculating the global motion for each grid map to the reference map. To solve this problem, many authors declaimed that their pair-wise merging approaches can be directly extended to merge multiple grid maps sequentially. More specifically, the pair-wise merging algorithm can repeatedly merge two grid maps and integrate them into one grid until all the grid maps are integrated together. However, this kind of approach suffers from the error accumulative problem. As mentioned in \cite{Zhu13,Ma16}, the problem of pair-wise grid map merging can be viewed as the pair-wise registration problem \cite{Besl92,Zhu114}. Accordingly, the problem of multiple grid map merging can also be viewed as multi-view registration problem \cite{Huber03,Ajmal06,Zhu16,Evang14,Govindu14,Zhu14,Fed16}. However, most of multi-view registration should be provided with the good initial motions in advance \cite{Evang14,Govindu14,Zhu14,Fed16}. Otherwise, they are unable to accomplish the multi-view registration. Besides, although some existing approaches can achieve multi-view registration without initial motions, they are designed to deal with 3D range scan and always time-consuming \cite{Huber03,Ajmal06,Zhu16}. Therefore, it is required to design an automatic multi-view registration approach, which can efficiently deal with 2D grid maps. Recently, motion averaging algorithm has been introduced as an effective means to solve the multi-view registration problem \cite{Govindu04}. Although this approach can effectively accomplish the multi-view registration, it should be provided with good initial global motions and reliable pair-wise registration results \cite{Govindu14,Govindu06}.

Based on the original motion averaging algorithm, this paper proposes an effective grid map merging approach, which can simultaneously merge multiple grid maps without any prior information. As it is difficult to directly calculate the global motions for these grid maps, the proposed approach accomplish the merging of multiple grid maps by three steps. Firstly, the pair-wise merging method is presented to estimate relative motions for the grid map pair, which has a certain amount of overlapping percentage. As the pair-wise merging algorithm may be applied to some grid map pairs, which have low overlapping percentages or even non-overlapping, the estimated relative motion may be unreliable. Therefore, all grid maps and the estimated relative motions are utilized to construct a undirected graph so as to sample the maximal connected subgraph (MCS). By confirming the sampled MCS with all relative motions, it is easy to calculate the initial global motions and eliminate all unreliable relative motions. Subsequently, the motion averaging algorithm can be adopted to refine the initial global motions so as to obtain accurate global motions for merging multiple grid maps. To illustrate its superiority, the proposed approach is tested on some real robot data sets.

This paper is organized as follows. In the next section, the grid map merging problem is stated and the TrICP algorithm is briefly reviewed. Section 3 proposes our approach for simultaneous merging of multiple grid maps. In Section 4, the proposed approach is tested and evaluated on three real robot data sets. Finally,some conclusions are drawn in Section 5.

\section{Problem Statement and the TrICP algorithm}
\label{sec:1}
This section firstly states the problem of grid map merging. As pair-wise map merging is the basis of multiple map merging, it then briefly reviews the 2D TrICP algorithm for the pair-wise map merging.

\subsection{Problem Statement}
\label{sec:2}

To build large grid map, mapping can be cooperatively implemented by multiple robots exploring different parts of the environment. Accordingly, a set of local grid maps built by different robots should be integrated into one global grid map.

Suppose there are two local grid maps built by robots exploring two parts of the same environment. According to \cite{Carpin08}, the goal of pair-wise map merging is to find a relative motion:
\begin{equation}
{\bf{M}} = \left[ {\begin{array}{*{20}{c}}
   {\bf{R}} & {t}  \\
   0 & 1  \\
\end{array}} \right],
\end{equation}
with which these two local maps can be properly integrated into a global map. More specifically, ${\bf{R}} \in \mathbb{R}^{2\times 2}$ denotes a rotation matrix determined by the angle $\theta$ and $\vec t \in {\mathbb{R}^2}$ is a translation vector:
\begin{equation}
{\bf{R}} = \left[ {\begin{array}{*{20}{c}}
   {\cos \theta } & { - \sin \theta }  \\
   {\sin \theta } & {\cos \theta }  \\
\end{array}} \right],\quad t = \left[ {\begin{array}{*{20}{c}}
   {{ t_x}}  \\
   {{ t_y}}  \\
\end{array}} \right].
\end{equation}

\begin{figure}
\begin{center}
\includegraphics[width= 0.45\linewidth]{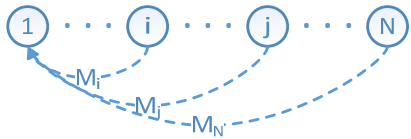}
\end{center}
   \caption{The diagram of multiple grid map merging, where one circle represents a grid map and dashed curves indicate global motions required to be estimated.}
\label{fig:Multi}
\end{figure}

Given a set of local grid maps, the goal of multiple grid map merging is to integrate these local maps into a single global map. Without loss of generality, the first grid map can be viewed as the reference map. As shown in Fig \ref{fig:Multi}, this merging problem is equivalent to calculating a set of global motions ${{\bf{M}}_{global}} = \{ {{\bf{I}}},{{\bf{M}}_2},...,{{\bf{M}}_N}\}$, so that these local maps can be properly merged into a global global map.

\subsection{The TrICP algorithm}
\label{sec:2}
Suppose there are two grid maps with non-overlapping areas, the model map $P$ and the subject map $Q$, where $\xi$ represents their overlapping percentage. By applying the edge extraction algorithm, two edge point sets $P \buildrel \Delta \over = \{ p_i\} _{i = 1}^{{N_p}}$ and $Q \buildrel \Delta \over = \{ q_j\} _{j = 1}^{{N_q}}$ can be extracted from these two grid maps to be merged. Denote ${P_\xi }$ as the point subset, which corresponds the overlapping part of the subject map to the model map. For pair-wise map merging, the relative motion ${\bf{M}}$ can be estimated by minimizing the following objective function:
\begin{equation}
\begin{array}{l}
 \mathop {\arg \min }\limits_{\xi ,{\bf{R}}, t} \frac{{\sum\limits_{{p_i} \in {P_\xi }} {\left\| {{\bf{R}}{{p}_i} + t - { q_{c(i)}}} \right\|_2^2} }}{{\left| {{P_\xi }} \right|{\xi ^{1 + \lambda }}}} \\
 {\rm{s}}{\rm{.t}}{\rm{.}}\quad \quad {{\bf{R}}^{\rm{T}}}{\bf{R}}{\rm{ = }}{{\rm{I}}_2},\det ({\bf{R}}) = 1 \\
 \end{array}
 \label{eq:TrICP}
\end{equation}
where ${{\rm{I}}_2}$ denotes the 2D identity matrix, $\lambda$ is a preset parameter and $\left|  \cdot  \right|$ indicate the cardinality of a set.

Actually, Eq. (\ref{eq:TrICP}) can be solved by the TrICP algorithm \cite{Chet05,Phils07}, which can obtain the optimal relative motion by iterations. Given the initial relative motion ${{\bf{M}}_0}$, three steps are included in each iteration of this algorithm:

(1)	Based on the previous motion, establish the point correspondence for each edge point in the subject map:
\begin{equation}
{c_k}(i) = \mathop {\arg \min }\limits_{j \in \{ 1,2, \cdots ,{N_q}\} } {\left\| {{{\bf{R}}_{k - 1}}{p_i} + {t_i} - {q_j}} \right\|_2}\quad \quad i = 1,2, \cdots {N_p}.
\end{equation}

(2)	Update the $k$th overlapping percentage and its corresponding subset:
\begin{equation}
({\xi _k},{P_{{\xi _k}}}) = \mathop {\arg \min }\limits_\xi  \sum\limits_{{p_i} \in {P_\xi }} {\left\| {{{\bf{R}}_{k - 1}}{p_i} + {t_{k - 1}} - {q_{{c_k}(i)}}} \right\|_2^2} /(\left| {{P_\xi }} \right|{(\xi )^{1 + \lambda }})
\end{equation}

(3)	Calculate the current relative motion:
\begin{equation}
{{\bf{M}}_k} \buildrel \Delta \over = ({{\bf{R}}_k},{t_k})\mathop { = \arg \min }\limits_{{\bf{R}},t} \sum\limits_{{p_i} \in {P_{{\xi _k}}}} {\left\| {{\bf{R}}{p_i} + t - { q_{c(i)}}} \right\|_2^2}
\label{eq:SVD}
\end{equation}

Finally, the optimal relative motion can be obtained by repeating these three steps until some stop conditions are satisfied. It should be noted that the TrICP algorithm can only obtain reliable relative motions for the grid map pair, which contains a certain amount of overlapping percentage \cite{Zhu114}.

\section{Merging multiple grid maps}

This section proposes the effective approach for simultaneous merging of multiple grid maps by the robust motion averaging.

\begin{figure}
\begin{center}
\includegraphics[width= 1.0\linewidth]{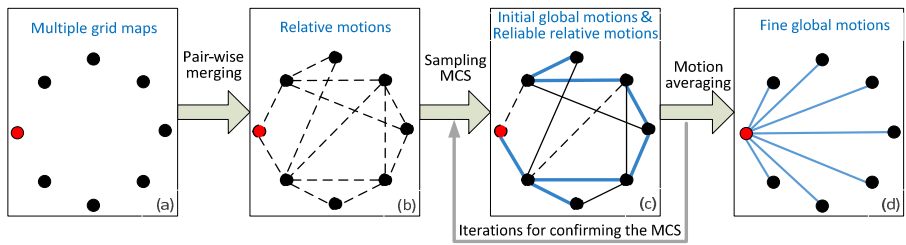}
\end{center}
   \caption{The flowchart of the proposed approach, where each vertex of graph denotes a grid map, each edge indicates the relative motion of its connected vertexes. (a) Multiple grid maps to be merged, where the red one denotes the reference map. (b) A set of relative motions are obtained from the pair-wise merging. (c) The confirmed MCS is connected by thick lines, which represents all reliable relative motions along with other solid lines. (d) The accurate global motions denoted by thick lines.}
\label{fig:Flow}
\end{figure}

Given a set of gird maps, the proposed approach can accomplish grid map merging by three steps displayed in Fig. \ref{fig:Flow}. Firstly, the pair-wise merging method is presented to estimate the relative motions for many grid map pairs. Subsequently, all grid maps and the estimated relative motions can be viewed as an undirected graph, where each vertex denotes a grid map and each edge indicates an estimated relative motion between the two vertices. Then, a randomized sampling scheme is utilized to find the maximal connected subgraph (MCS). As there may exist unreliable relative motions obtained from the pair-wise merging step, the sampling MCS should be confirmed by all relative motions. The process of MCS sampling and confirming should be repeated until the preset number of iterations so as to search for the optimal MCS and eliminate unreliable relative motions. Finally, the accurate global motions can be recovered by the application of the 2D motion averaging algorithm to all reliable relative motions.

\subsection{Pair-wise grid map merging}

To estimate the relative motion ${{\bf{M}}_{ij}}$,  the pair-wise grid map merging method should be well designed. As mentioned before, the TrICP algorithm can be utilized to estimate the relative motion of one map pair which includes a certain amount of overlapping percentage. However, owing to the local convergence property, good initial relative motion should be provided to the TrICP algorithm. Otherwise, it is easy to be trapped into the local minimum and obtain the unreliable relative motion.

\begin{figure}
\begin{center}
\includegraphics[width= 0.6\linewidth]{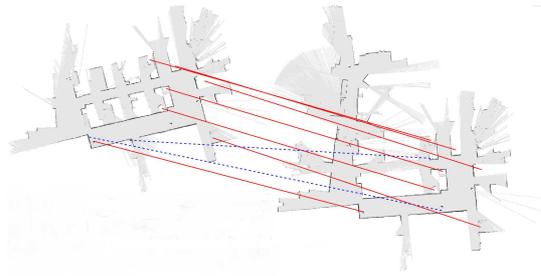}
\end{center}
   \caption{SIFT features extracted and matched from grid map pair, where solid red lines denote the true matches and dashed blue lines indicate the false matches.}
\label{fig:SIFT}
\end{figure}

For the pair-wise map merging, the scale-invariant feature transform (SIFT) futures \cite{Lowe04,Brown07} can be extracted from two grid maps respectively. As the SIFT features are invariant to rotation and translation changes, it is easy to establish feature matches between these two grid maps. Due to the sensor noise and the accuracy of mapping algorithm, there might exist some false matches. As shown in Fig. \ref{fig:SIFT}, there are two grid maps $P$ and $Q$, which include overlapping areas. Suppose there are a set of SIFT feature matches $\{ {F_{i,P}},{F_{i,Q}}\} _{i = 1}^N$, which are extracted and matched from these two grid maps. Obviously, if the match $\{ {F_{i,P}},{F_{i,Q}}\}$ is true, the SIFT features ${F_{i,P}}$ and ${F_{i,Q}}$ must correspond to the same location of the environment, and they should satisfy the following equation:
\begin{equation}
\left\| {{\bf{R}}{f_{i,P}} + t - {f_{i,Q}}} \right\|_2^2 \approx 0,
\label{eq:Cons}
\end{equation}
where ${\bf{M}} \buildrel \Delta \over = ({\bf{R}}, t)$ denotes the relative motion of these two grid maps, ${f_{i,P}}$ and ${f_{i,Q}}$ represent the locations of SIFT features ${F_{i,P}}$ and ${F_{i,Q}}$, respectively. However, the false feature match does not meet this requirement.

According to Eq. (\ref{eq:SVD}), two true feature matches are enough to estimate the initial relative motion for the TrICP algorithm. Therefore, the random sample consensus (RANSAC) algorithm can be used to find the true matches. More specifically, two feature matches can be randomly selected from all feature matches so as to calculate the guess of relative motion ${\bf{\tilde M}}$, then Eq. (\ref{eq:Cons}) can be used to test all established feature matches and count the number of true feature matches. And the best guess ${{\bf{\tilde M}}_{best}}$ corresponds to the one, which can receive the support of all true matches. To obtain the best guess, the random guess should be repeatedly generated and tested until the preset maximum number of iteration reaches. Finally, the best guess ${{\bf{\tilde M}}_{best}}$ can be viewed as the initial relative motion of the TrICP algorithm so as to refine the relative motion of two grid maps to be merged.

Based on the above description, the proposed pair-wise map merging method can be summarized as the Algorithm 1.
\begin{algorithm}[htbp]
        \caption{: Pair-wise grid map merging algorithm}
               \textbf{Input}: Grid maps  $P$ and $Q$

                 \textbf{Output}: Estimation of the relative motion ${\bf{\hat M}}$

            \quad Extract SIFT features for  $P$ and $Q$, respectively;

            \quad Establish all the feature matches $\{ {F_{i,P}},{F_{i,Q}}\} _{i = 1}^N$
              and set $k$ = 0;

       \quad \textbf{Do}

               \qquad $k = k + 1$;

               \qquad Randomly select two matches ${\{ {F_{i,P}},{F_{i,Q}}\} _{i = m,n}}$;

               \qquad Calculate the motion guess ${{\bf{\tilde M}}_k}$ by Eq.(6);

               \qquad Compute ${d_i}= \left\| {{\bf{R}}{{f}_{i,P}} + t - {{f}_{i,Q}}} \right\|_2$ for each feature match;

               \qquad Count the number ${N_k}$ of feature matches with ${d_i} \le {d_{thr}}$;

               \qquad \textbf{If} ${N_k} > {N_{best}}$

               \qquad \quad ${N_{best}} = {N_k}$;

               \qquad \quad  ${{\bf{\tilde M}}_{best}} = {{\bf{\tilde M}}_k}$;

               \qquad \textbf{End}

           \quad \textbf{While} ($k < 200$)

          \quad Extract the edge point sets $P \buildrel \Delta \over = \{ {p_i}\} _{i = 1}^{{N_p}}$ and $Q \buildrel \Delta \over = \{ {q_j}\} _{j = 1}^{{N_q}}$;

          \quad Obtain ${\bf{\hat M}}$ by refining  ${{\bf{\tilde M}}_{best}}$ with the TrICP algorithm.

    \end{algorithm}
Theoretically, two true feature matches are enough to estimate the initial relative motion for the TrICP algorithm. However, if the number of true matches is less than three, there is no way to confirm and calculate the correct initial motion. To guarantee the robustness, the TrICP algorithm is only applied to these map pairs, which satisfy ${{\bf{\tilde M}}_{best}} \ge 4$. Otherwise, there is no need to apply the TrICP algorithm. Suppose SIFT features has been extracted for grid maps $P$ and $Q$. To establish the feature matches, we can either search the nearest neighbor from the map $Q$ for each SIFT feature in the map $P$ or vice verse. In practice, these two strategies can obtain different number of consistent matches for these two grid maps to be merged. Therefore, during the establishment of feature matches, both strategies should be implemented so as to obtain as many consistent matches as possible.

After the application of pair-wise map merging, a set of relative motions can be obtained for the construction of undirected graph so as to sample and confirm the optimal MCS.

\subsection{MCS sampling and confirming}

Among these estimated relative motions, there may exist some unreliable relative motions due to the unreasonable application of the pair-wise merging method to these grid map pairs, which contain low percentage or even non-overlapping. Therefore, the optimal MCS should be confirmed so as to calculate initial global motions and eliminate unreliable relative motions for the motion averaging.

\begin{figure}
\begin{center}
\includegraphics[width= 0.25\linewidth]{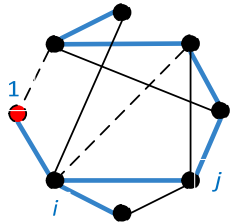}
\end{center}
   \caption{Diagram of the sampled MCS, which is connected by thick lines. Each dashed line denotes one unreliable relative motion.}
\label{fig:MCS}
\end{figure}

Given a set of relative motions ${\rm{\{ \hat M}}_{ij}^r{\rm{\} }}_{r = 1}^R$, it is easy to construct an undirected graph $G$, where one vertex denotes a grid map and each edge indicates the estimated relative motion of its connected grid maps. Accordingly, global motions can be estimated from the MCS, which is composed of $(N-1)$ edges and $N$ vertexes of the graph $G$. As displayed in Fig. \ref{fig:MCS}, based on the MCS, the global motion guess of the $i$th grid map can be directly set as ${{\bf{\tilde M}}_i} = {{\bf{\hat M}}_{1i}}$, where ${{\bf{\hat M}}_{1i}}$ has been estimated by the pair-wise map merging. Subsequently, the global motion of the $j$th grid map can be calculated as:
\begin{equation}
    {{\bf{\tilde M}}_j} = {{\bf{\tilde M}}_i}{{\bf{\hat M}}_{ij}}.
    \label{eq:Mj}
\end{equation}
where ${{\bf{\hat M}}_{ij}}$ has been estimated and included in the relative motion set ${\rm{\{ \hat M}}_{ij}^r{\rm{\} }}_{r = 1}^R$. As the MCS exits a path between the 1st vertex to all other vertexes in the $G$, Eq. (\ref{eq:Mj}) can be transitively used to calculate all other global motions. The main questions arising here are how to sample the MCS from the graph $G$ and how to confirm the optimal MCS.

To sample a MCS, we can set a null matrix ${\bf{L}}$ of the size $N \times N$. As one MCS contains $(N-1)$ edges of the graph $G$, a subgraph ${G}'$ with all vertex of $G$ can be generated by the random selection of $(N-1)$ relative motions from the motion set ${\rm{\{ \hat M}}_{ij}^r{\rm{\} }}_{r = 1}^R$. Then we can set ${\bf{L}}(i,j) = 1$, if the corresponding relative motion ${{\rm{\hat M}}_{ij}}$ is included in the subgraph ${G}'$. Subsequently, a matrix ${\bf{g}}$ can be calculated as follows:
\begin{equation}
    {\bf{g}} = {({\bf{L}} + {\bf{L}'} + {{\bf{I}}_N})^N}
    \label{eq:MCS}
\end{equation}
where ${{\bf{I}}_N}$ denotes the identity matrix of the size $N \times N$. If and only if all the elements of the matrix ${\bf{g}}$ are non-zeros, the subgraph ${G}'$ can be viewed as a MCS of the graph $G$.

As displayed in Fig. \ref{fig:MCS}, only $(N-1)$ relative motions are contained in the sampled MCS. Hence, all other relative motions can be used to confirm the sampled MCS. Because each edge of the optimal MCS corresponds to a reliable relative motion, Eq.(\ref{eq:Mj}) can be transitively used to calculated all global motions  ${{\bf{\tilde M}}_{global}} = \left\{ {{\bf{I}},{{{\bf{\tilde M}}}_2},...,{{{\bf{\tilde M}}}_m},...,{{{\bf{\tilde M}}}_n},{{{\bf{\tilde M}}}_N}} \right\}$ with good accuracy. Suppose the graph G includes an reliable relative motion ${{\bf{\hat M}}_{mn}}$, which is not contained in the optimal MCS. Since the relative motion ${{\bf{\hat M}}_{mn}}$ estimated by the pair-wise merging algorithm, it inevitably contains error. Therefore,
\begin{equation}
    {{\bf{\hat M}}_{mn}} \approx {{\bf{\tilde M}}_m}^{ - 1}{{\bf{\tilde M}}_n}.
\end{equation}

However, this relationship no longer holds for the unreliable relative motions. In practice, Eq. (10) can be replaced by the following constraint:
\begin{equation}
     d({{\bf{\hat M}}_{ij}},{{\bf{\tilde M}}_i}^{ - 1}{{\bf{\tilde M}}_j}) = {\left\| {{{{\bf{\hat M}}}_{ij}} - {{\bf{\tilde M}}_i}^{ - 1}{{{\bf{\tilde M}}}_j}} \right\|_F} \le {d_{thr}}
\end{equation}

where ${d_{thr}}$ denotes the preset distance threshold. Based on this constraint, all estimated relative motions can be used to confirm the optimal MCS, which can receive the support of most relative motions in the graph.

The randomly sampled MCS is not necessary optimal due to the existence of unreliable relative motions, so the sampling and confirming of MCS should be repeatedly until the preset maximum number of iterations are reached. Accordingly, the proposed MCS sampling and confirming method can be summarized as the Algorithm 2.

\begin{algorithm}[htbp]
        \caption{: MCS sampling and confirming}
               \textbf{Input}: All the relative motions $\{ {{\bf{\hat M}}_{ij}^r} \}_r^R$

                 \textbf{Output}: Global motions ${{\bf{\hat M}}_{global}}$ and reliable relative motions $\{ {{\bf{\hat M}}_{ij}^r} \}_r^{R'}$

            \quad ${E_{best}} = 0$ and $k=0$;

            \quad Construct the graph G based on $\{ {{\bf{\hat M}}_{ij}^r} \}_r^{{R}}$;

       \quad \textbf{If} $(k \le 10{N^2})$

               \qquad $k= k+ 1$;

               \qquad \textbf{do}

               \qquad \quad Sample the subgraph ${{\bf{G}}'}$ from the graph $G$;

               \qquad \quad Compute the matrix ${\bf{g}}$ denoted by Eq. (\ref{eq:MCS});

               \qquad \textbf{Until} (All elements of ${\bf{g}}$ are non-zeros)

               \qquad Estimate ${\bf{\tilde M}}_{global}^r$ from the MCS by Eq. (\ref{eq:Mj});

                \qquad Count the number ${E_r}$ of edges that satisfy $d({{\bf{\hat M}}_{ij}},{{\bf{\tilde M}}_i}^{ - 1}{{\bf{\tilde M}}_j}) \le {d_{thr}}$;

                \qquad \textbf{If} (${E_r} \ge {E_{best}}$)

                \qquad \quad ${{\bf{\hat M}}_{global}} = {\bf{\tilde M}}_{global}^r$

                \qquad \quad Eliminate edges from $\{ {{\bf{\hat M}}_{ij}^r} \}_r^R$, which satisfy $d({{\bf{\hat M}}_{ij}},{{\bf{\tilde M}}_i}^{ - 1}{{\bf{\tilde M}}_j}) > {d_{thr}}$;

                \qquad \textbf{end}

       \quad \textbf{end}

    \end{algorithm}

After the application of MCS sampling and confirming, the initial global motions and a set of reliable relative motions can be obtained for the motion averaging.

\subsection{Motion Averaging}

Although global motions have been estimated from the optimal MCS by transitively using Eq. (\ref{eq:Mj}),
they are coarse due to the accumulative error. Since a set of reliable relative motions have been confirmed by
the optimal MCS, they can be incorporated to optimize the coarse global motions.
The key question arising here is how to use these 2D relative motions so as to refine the coarse global motions. In \cite{Govindu04}, Govindu $et$ $al$. proposed the 3D motion averaging algorithm, which can refine the coarse global motions by a set of relative motions. For the 2D motion, the original motion averaging algorithm should be extended.

In fact, the 2D motion ${\bf{M}} \in SE(2)$ belongs to the Lie group and its logarithm ${\bf{M}}$ belongs to the Lie algebra ${\bf{m}} \in SE(2)$, which can be denoted as follows:
\begin{equation}
{\bf{m}} = {\mathop{\rm logm}\nolimits} ({\bf{M}}) = \left[ {\begin{array}{*{20}{c}}
   {\bf{\Omega}}  & u  \\
   0 & 0  \\
\end{array}} \right],
\end{equation}
where $ u = {[{u_1},{u_2}]^T}$ is a vector and ${\bf{\Omega}}$ is a skew-symmetric matrix:
\begin{equation}
 {\vec{\Omega}} {\rm{ = }}\left[ {\begin{array}{*{20}{c}}
   1 & {{{\vec{\Omega}} _{12}}}  \\
   { - {{\vec{\Omega}} _{12}}} & 1  \\
\end{array}} \right].
\end{equation}
Accordingly, the Lie algebra ${\vec{m}} \in SE(2)$ can be transformed into other form ${v} = vec({\bf{m}})$, where $vec(.)$ indicates the function which can arrange all parameters of ${\bf{m}}$ into a compated 3D column vector. Vice verse, $rvec(.)$ can be utilized to denote the inverse function of $vec(.)$.
By applying the first-order approximation to the Riemannian distance \cite{Govindu04}, there exists the following relationship for two approximate motions ${{\bf{M}}_i}$ and ${{\bf{M}}_j}$:
\begin{equation}
\begin{array}{l}
 {\rm{     logm({{\bf{M}}}}_i}^{ - 1}{{\rm{{\bf{M}}}}_j}{\rm{)}} \approx {\rm{logm(}}{{\rm{{\bf{M}}}}_i}{\rm{) - logm(}}{{\rm{{\bf{M}}}}_j}{\rm{)}} \\
  \Rightarrow {{\bf{m}}_{ij}} \simeq {{\bf{m}}_i}{\bf{ - }}{{\bf{m}}_j}, \\
 \end{array}
 \label{eq:Lie}
\end{equation}
where the more these two motions are approximate, the more ${{\bf{m}}_{ij}}$ approximates to the term $({{\bf{m}}_i}{\bf{ - }}{{\bf{m}}_j})$.

Suppose ${{\bf{M}}_i}(\bf{M_j})$ denotes the global motion of the $i$th ($j$th) grid map to the reference map, ${{\bf{M}}_{ij}}$ indicates the relative motion between the $i$th grid map and the $j$th grid map. They obey the constraint ${{\bf{M}}_{ij}} = {{\bf{M}}_i}^{ - 1}{{\bf{M}}_j}$ . For the problem of multiple map merging, the motions ${{\bf{M}}_i}$ and ${{\bf{M}}_j}$ are variables required to be estimated. While, ${{\bf{M}}_{ij}}$ can be approximated by the one ${{\bf{\hat M}}_{ij}}$ estimated from the pair-wise map merging. In other words, ${{\bf{M}}_{ij}}$ and ${{\bf{\hat M}}_{ij}}$ is very approximate. Therefore:
\begin{equation}
\Delta {{\bf{m}}_{ij}} = {\rm{logm}}({{\bf{M}}_i}{{\bf{\hat M}}_{ij}}{{\bf{M}}_j}^{ - 1}) = \Delta {{\bf{m}}_j} - \Delta {{\bf{m}}_i}.
\end{equation}
As the column vector $v$ represents another form of {\bf{m}}, the same relationship also holds for the column vector, i.e. $\Delta {{v}_{ij}} = \Delta {{v}_j} - \Delta {{v}_i}$. Obviously, all the column vectors $\{ \Delta {{v}_i}\} _{i = 1}^N$ can be concatenated into one large vector $\Im {\rm{ = [}}{{\rm{{v}}}_1};{{v}_2}; \cdots ;{{v}_N}]$. Subsequently, the equation $\Delta {{v}_{ij}} = \Delta {{v}_j} - \Delta {{v}_i}$ can be transformed into the following form:
\begin{equation}
\Delta {v_{ij}} = {{\bf{D}}_{ij}}\Im  = [ \cdots ,{{\bf{I}}_3}, \cdots , - {{\bf{I}}_3}]\Im
\label{eq:ma}
\end{equation}
where ${{\bf{I}}_3}$ is the 3D identity matrix, ${{\bf{D}}_{ij}}$ can be viewed as an indicator matrices of size $3 \times (3N - 3)$ with matrices ${{\bf{I}}_3}$ and $ - {{\bf{I}}_3}$ at position $j$ and $i$, respectively. As there are a set of reliable relative motions confirmed by the optimal MCS, it is convenient to concatenate all increment vectors of relative motions into one large vector ${\bf{V}} = \left[ {\begin{array}{*{20}{c}}
   {\Delta {{v}_{i{j_1}}};} & {\Delta {{v}_{ij2}};} & {...}  \\
\end{array}} \right]$, Similarly, all the indicator matrices can also be concatenated into one large matrix ${\bf{D}} = \left[ {\begin{array}{*{20}{c}}
   {{{\bf{D}}_{i{j_1}}};} & {{{\bf{D}}_{ij2}};} & {...}  \\
\end{array}} \right]$. According to Eq. (\ref{eq:ma}), there exists the following relationships:
\begin{equation}
{\bf{V}} = {\bf{D}}\Im
\end{equation}
and
\begin{equation}
\Im  = {{\bf{D}}^\dag }{\bf{V}},
\end{equation}
where ${{\bf{D}}^\dag }$ denotes the pseudo inverse matrix of ${\bf{D}}$. Given the initial global motion $\{ {{\bf{\hat M}}_i}\} _{i = 1}^N$, the increment vectors $\{\Delta {{v}_i}\} _{i = 1}^N$ can be incorporated to refine the global motion as follows:
\begin{equation}
{{\bf{M}}_i} = {\rm{expm(rvec(}}\Delta {{\vec{v}}_i}{\rm{))}}{{\bf{\hat M}}_i}\quad (i = 2,3,...,N)
\end{equation}
where the function $expm(.)$ denotes the exponential operation of matrix. As displayed in Eq. (\ref{eq:Lie}), the motion averaging algorithm cannot obtain the closed-form solution for global motions, so it is required to repeat the refinement until some stop conditions are satisfied. The sketch of the global motion refining algorithm is shown in Algorithm 3.
\begin{algorithm}[htbp]
        \caption{: Global motion refining}
               \textbf{Input}:    Initial global motions ${\bf\hat{M}}_{global} = \{ {\rm{I}},{{\bf{\hat M}}_2}, \cdots ,{{\bf{\hat M}}_N}\} $

                \qquad\qquad reliable relative motions $\{ {{\bf{\hat M}}_{ij}^r}\} _{r = 1}^{R'}$

                 \textbf{Output}: Fine global motions ${\bf{M}}_{global} = \{ {\rm{I}},{{\bf{M}}_2}, \cdots ,{{\bf{M}}_N}\} $

       \quad \textbf{Do}

               \qquad $\Delta {{\bf{M}}_{ij}} = {{\bf{\hat M}}_i}{{\bf{\hat M}}_{ij}}{{\bf{\hat M}}_j}^{ - 1}$;

               \qquad $\Delta {{\bf{m}}_{ij}} = \log (\Delta {{\bf{M}}_{ij}})$;

               \qquad $\Delta {{\rm{v}}_{ij}} = vec(\Delta {{\bf{m}}_{ij}})$;

               \qquad $\Im {\rm{ = }}{D^\dag }{V_{ij}}$;

              \qquad \textbf{for}

               \qquad \quad $\Delta {{\bf{m}}_i} = rvec(\Delta {v_i})$;

               \qquad \quad ${{\bf{M}}_i} = \exp (\Delta {{\bf{m}}_i}){{\bf{M}}_i}$;

               \qquad \quad ${{\bf{\hat M}}_i} ={{\bf{M}}_i}$;

               \qquad \textbf{end}

           \quad \textbf{Until} ${\rm{|}}\Delta \Im {\rm{|| < }}\varepsilon $

    \end{algorithm}

After the application of motion averaging, accurate global motions can be obtained for the merging of multiple grid maps.

\subsection{Implementation}

Given a set of unordered grid maps, the relative motions of grid maps can be estimated by the pair-wise map merging. As there may be exist unreliable relative motions, an undirected-graph can be constructed by all grid maps and their estimated relative motions. Accordingly, the MCS can be randomly sampled and then confirmed by all estimated relative motions.
By repeating the process of MCS sampling, the optimal MCS can be confirmed to calculate the initial global motions and select all reliable relative motions. Consequently, the initial global motions can be refined by applying the motion averaging algorithm to all reliable relative motions. Based on the refined global motions, the set of grid maps can be integrated into a single global map. Therefore, the proposed approach can be outlined in Algorithm 4.

 \begin{algorithm}[htbp]
        \caption{: Simultaneous merging multiple grid maps}
               \textbf{Input}: A set of unordered grid maps

                 \textbf{Output}: Merged map

               \qquad Extract the SIFT features and edge point sets from all grid maps;

               \qquad Estimate the relative motions for many map pairs by Algorithm 1;

               \qquad Obtain the initial global motions and reliable relative motions by the Algorithm 2;

               \qquad Acquire the fine global motions by the Algorithm 3;

               \qquad Merge all grid maps based on the fine global motions.

    \end{algorithm}

\section{Experimental Results}

\begin{figure}
\begin{center}
\includegraphics[width= 0.85\linewidth]{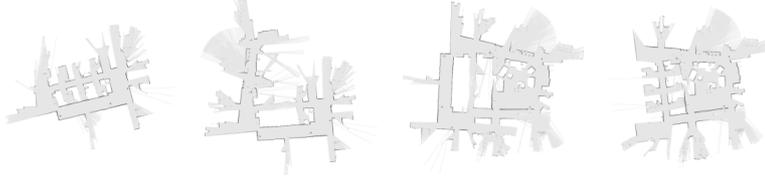}
\end{center}
   \caption{Grid maps built from Tim.log data set, which is divided into four parts.}
\label{fig:Tim}
\end{figure}

\begin{figure}
\begin{center}
\includegraphics[width= 0.85\linewidth]{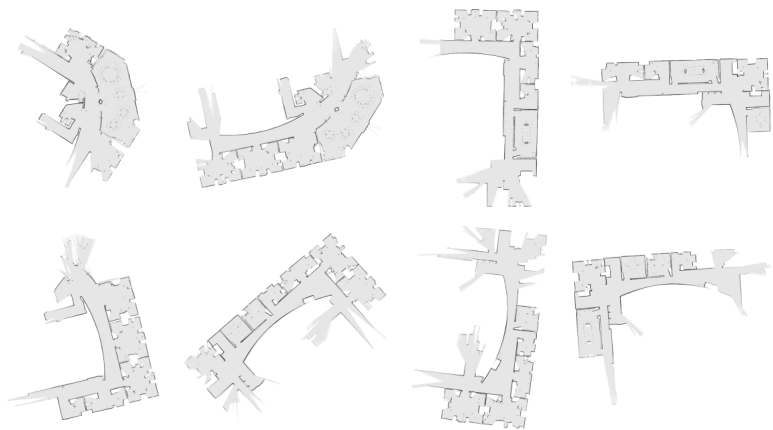}
\end{center}
   \caption{Grid maps built from Intel.log data set, which is divided into eight parts.}
\label{fig:Intel}
\end{figure}

To verify the performance of the proposed approach, a set of experiments were tested on three public datasets: Tim.log \cite{Bailey}, Intel.log \cite{Stachniss15} and Fr079.log \cite{Stachniss15}, which were recorded by mobile robots equipped with a laser range finder and odometer. All these datasets were recorded in door environment. To simulate multi-robot systems, these three data sets can be separated into four, eight and eleven parts, respectively. By applying the simultaneous localization and mapping(SLAM) algorithm \cite{Giorg07,Parr05}, they can be used to build grid map sets for testing the proposed approach. These grid map sets are displayed in Figs. \ref{fig:Tim}, \ref{fig:Intel} and \ref{fig:Fr079}
Experiments were implemented in MATLAB on a four-core 3.6GHz computer with 8GB of memory.

\subsection{Validation}

\begin{figure}
\begin{center}
\includegraphics[width= 0.85\linewidth]{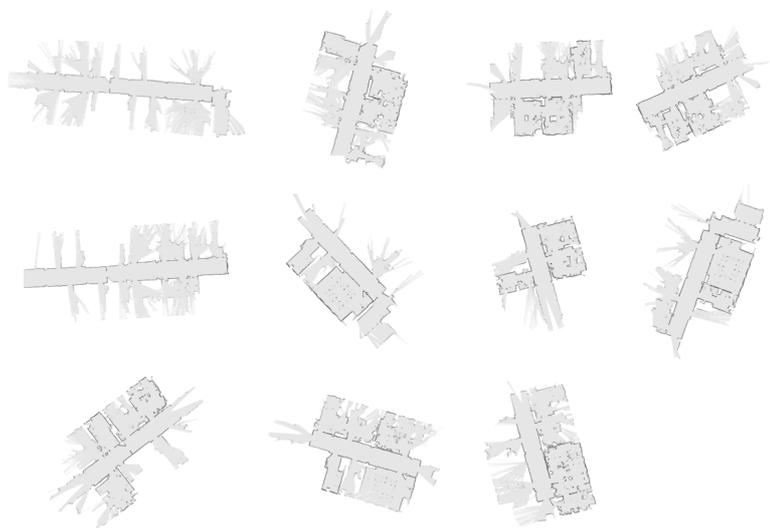}
\end{center}
   \caption{Grid maps built from Fr079.log data set, which is divided into eleven parts.}
\label{fig:Fr079}
\end{figure}

To validate the proposed approach, it was firstly tested on the grid map set built from Fr079.log. As shown in Fig. \ref{fig:Fr079}, there are eleven unordered grid maps, which require to be merged.


\begin{figure}[htp]
\begin{center}
\subfigure[]{\label{fig:IntMid-a}\includegraphics[scale=0.28]{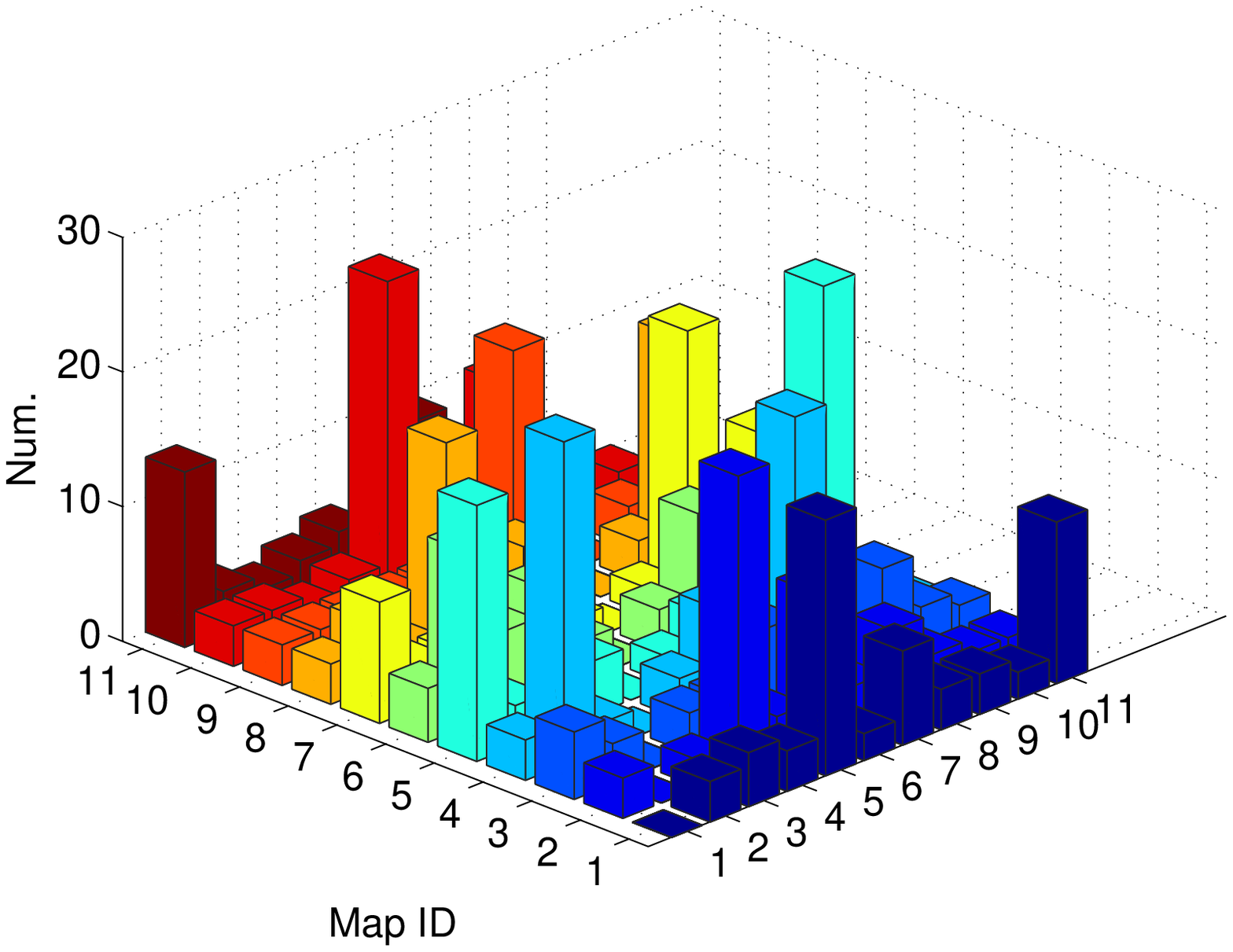}}
\subfigure[]{\label{fig:IntMid-b}\includegraphics[scale=0.28]{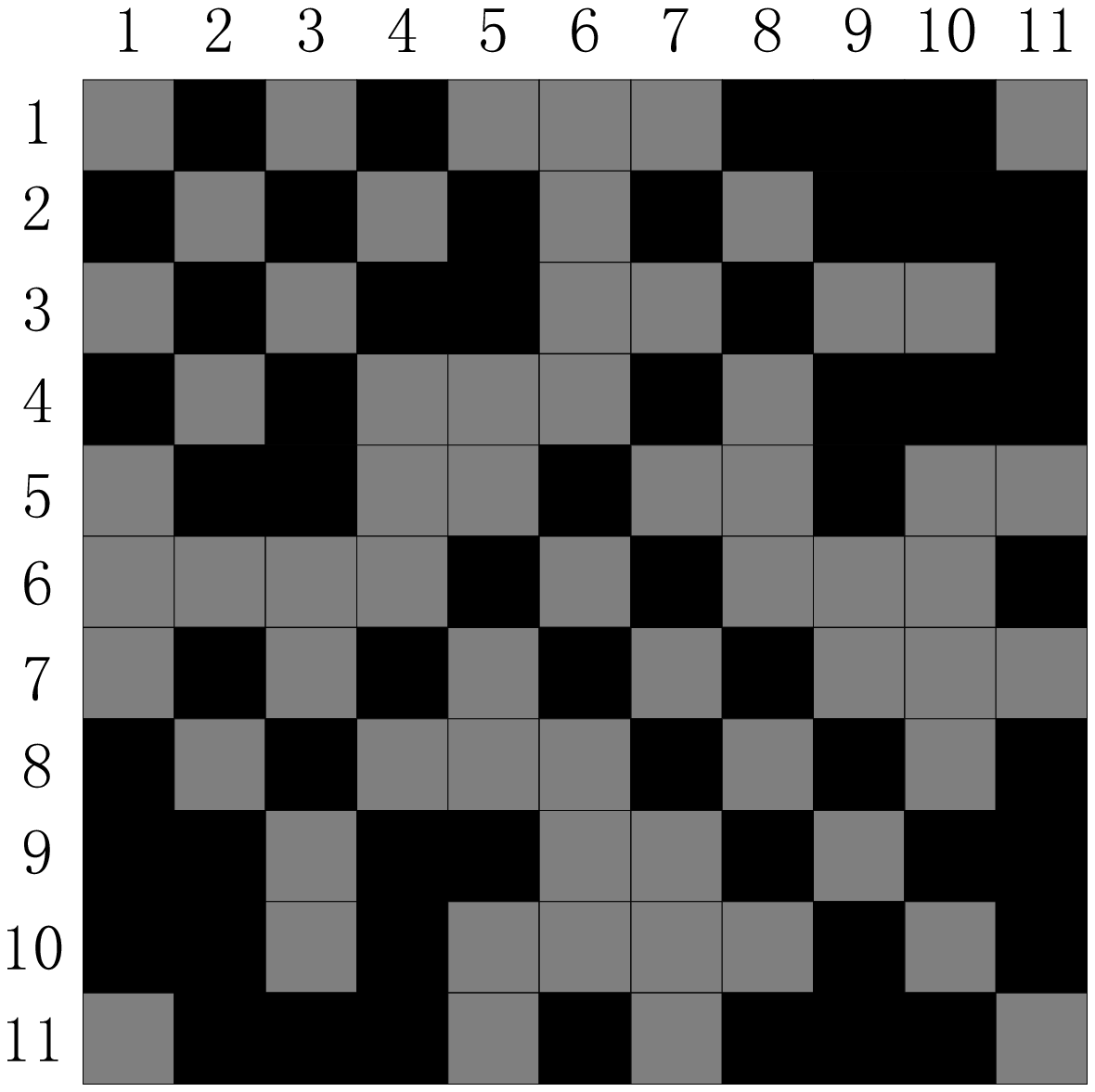}}
\subfigure[]{\label{fig:IntMid-c}\includegraphics[scale=0.28]{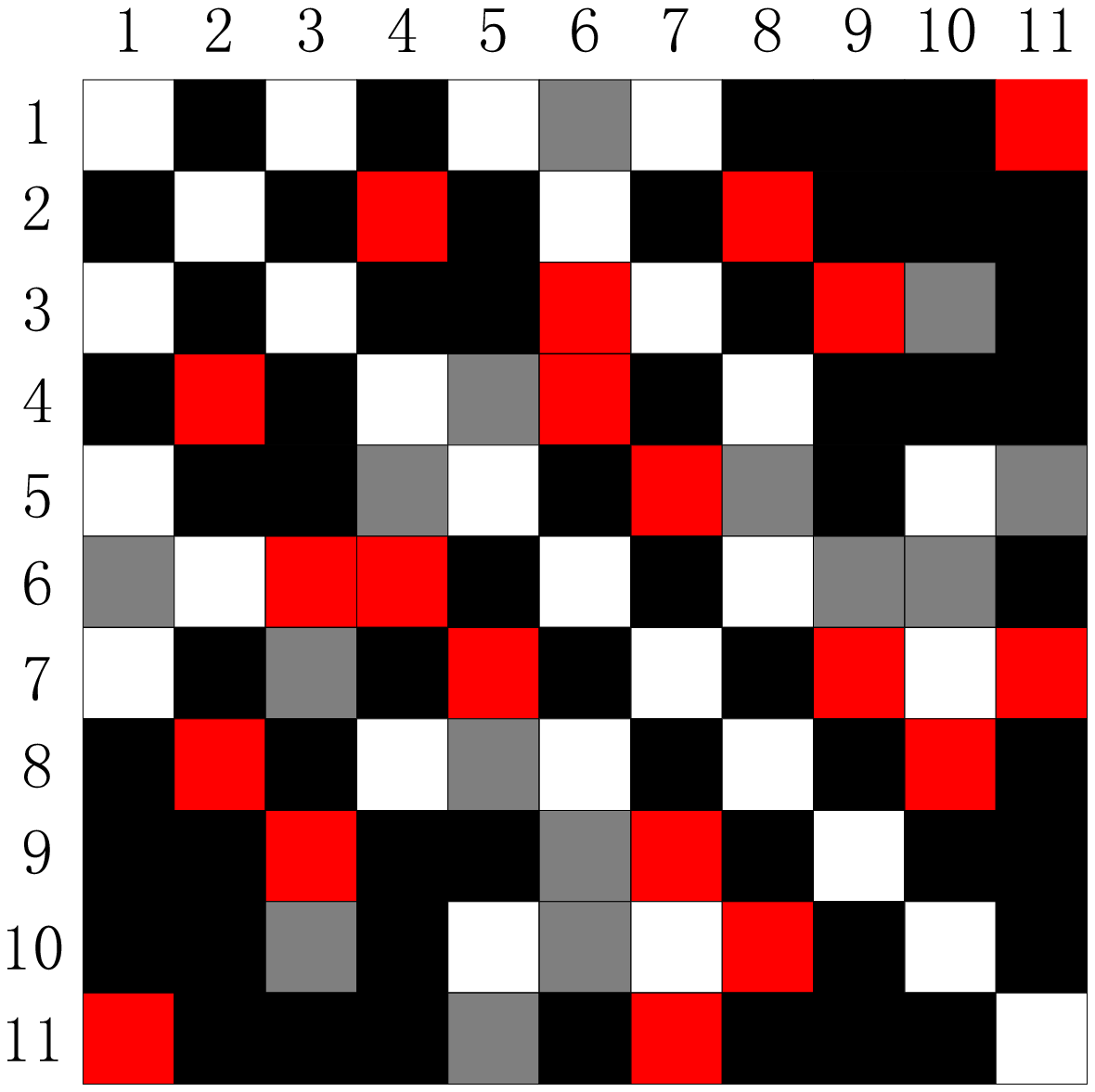}}
\end{center}
\caption{ Intermediate results of the proposed approach. (1) The number of detected true feature matches of each map pairs. (2) Map pairs containing estimated relative motions are denoted in gray. (3) The relative motions of the optimal MCS are denoted in the red, other reliable relative motions are denoted in the white and unreliable relative motions are denoted in the gray.}
\label{fig:IntMid}
\end{figure}

At the beginning, the pair-wise merging method should be utilized to calculate the relative motions of grid map pairs.
During pair-wise merging, true feature matches can be detected between each grid map pairs.
Fig. \ref{fig:IntMid-a} displays the detected number of true feature matches for all grid map pairs. As shown in Fig. \ref{fig:IntMid-a}, there are a portion of map pairs, which are lack of enough true feature matches due to the low overlapping percentages or even non-overlapping. For these map pairs, it is difficult to estimate their relative motions. For efficiency, the proposed approach only applies the pair-wise merging method to these map pairs, which at least contains four detected true feature matches. Given the true feature matches, initial relative motions can be provided to the TrICP algorithm so as to refine the relative motions for grid map pairs. Fig. \ref{fig:IntMid-b} indicates these map pairs, which can obtain their estimated relative motions. Due to some reasons, the pair-wise merging method may obtain some unreliable relative motions.

\begin{figure}[htp]
\begin{center}
\subfigure[]{\label{fig:ComFr-a}\includegraphics[scale=0.7]{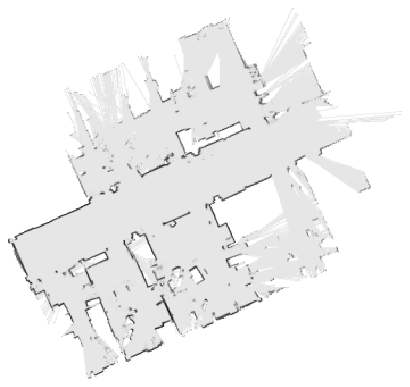}}
\subfigure[]{\label{fig:ComFr-b}\includegraphics[scale=0.7]{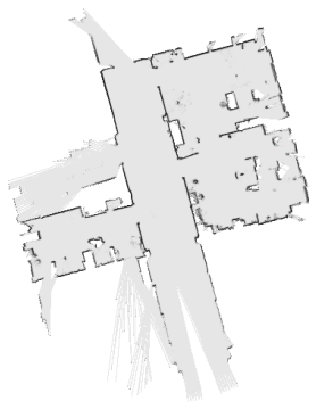}}
\subfigure[]{\label{fig:ComFr-a}\includegraphics[scale=0.7]{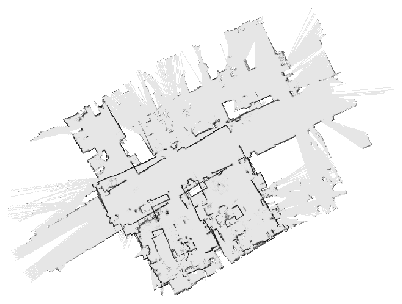}}
\subfigure[]{\label{fig:ComFr-b}\includegraphics[scale=0.7]{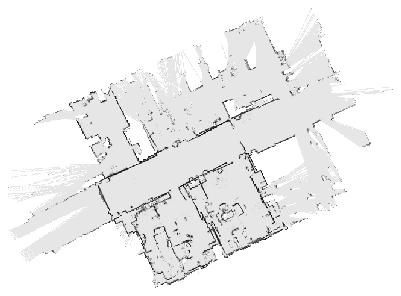}}
\end{center}
   \caption{The demonstration of one unreliable relative motion, which is estimated by the pair-wise merging method.
   (a) Model map. (b) Subject map. (c) Merging result based on the initial relative motions. (c) Merging result based on the estimated relative motions}
\label{fig:Unrm}
\end{figure}

Subsequently, the undirected graph should be constructed based on all grid maps and estimated relative motions.
On the constructed graph, it is easy to randomly sample the MCS, which contains $(N-1)$ estimated relative motions.
As the number of estimated relative motions are more than $(N-1)$, the residual relative motions can be utilized to confirm
whether the randomly sampled MCS is the optimal one or not. The process of sampling and confirming MCS should be repeated until the preset iteration number is reached. As a result, the optimal MCS can be searched out with all the reliable relative motions. Fig. \ref{fig:IntMid-c} displays all reliable relative motions and $(N-1)$ relative motions involved in the optimal MCS. As shown in \ref{fig:IntMid-c}, there are some of map pairs, whose estimated relative motions are unreliable. These unreliable relative motions may be caused by two reasons: (1) False true feature matches can only provide invalid initial relative motions to the TrICP algorithm. (2) Even given moderate initial relative motions, the TrICP algorithm may be trapped into local minimum due to the property of local convergence. To view them in a more intuitive way, Fig. \ref{fig:Unrm} displays the merging results of one map pair, which is denoted in the gray in \ref{fig:IntMid-c}.
As shown in Fig. \ref{fig:Unrm}, the relative motion of this map pair is really unexpected, so it should be eliminated by the optimal MCS.

As the optimal MCS contains the minimum set of good relative motions, they can be employed to estimate initial global motions. Fig. \ref{fig:ComFr-a} shows the multiple map merging results based on the initial global motions. As shown in Fig. \ref{fig:ComFr-a}, the initial global motions are not so satisfactory due to the accumulative errors. Hence, they should further be refined by the motion averaging algorithm. With all reliable relative motions, the motion averaging algorithm can calculate accurate global motions for the merging of multiple grid maps. Fig. \ref{fig:ComFr-b} illustrates the final merging result of multiple grid maps. As shown in Fig. \ref{fig:ComFr}, it is really necessary to apply the motion averaging algorithm, which can result in good merging results.

\begin{figure}[htp]
\begin{center}
\subfigure[]{\label{fig:ComFr-a}\includegraphics[scale=0.48]{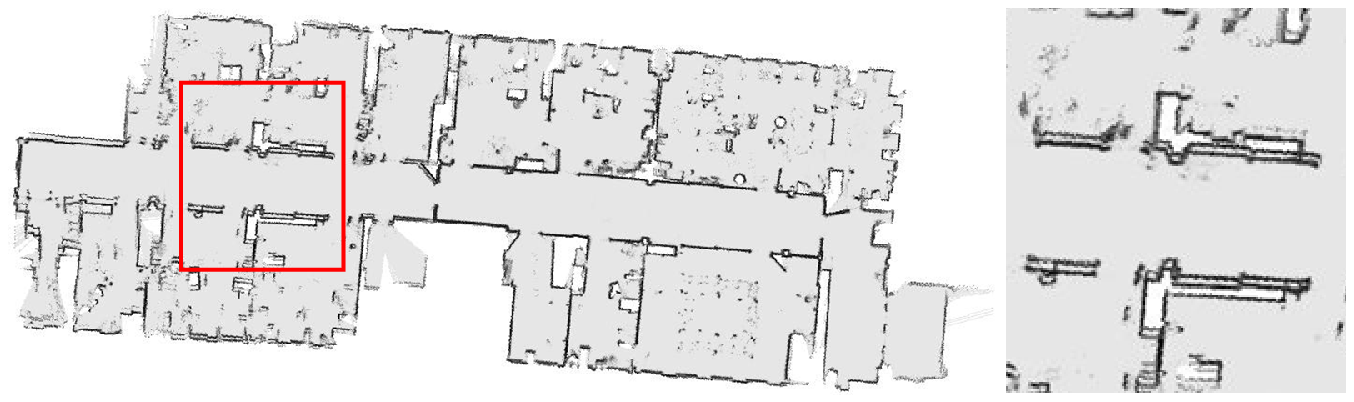}}
\subfigure[]{\label{fig:ComFr-b}\includegraphics[scale=0.48]{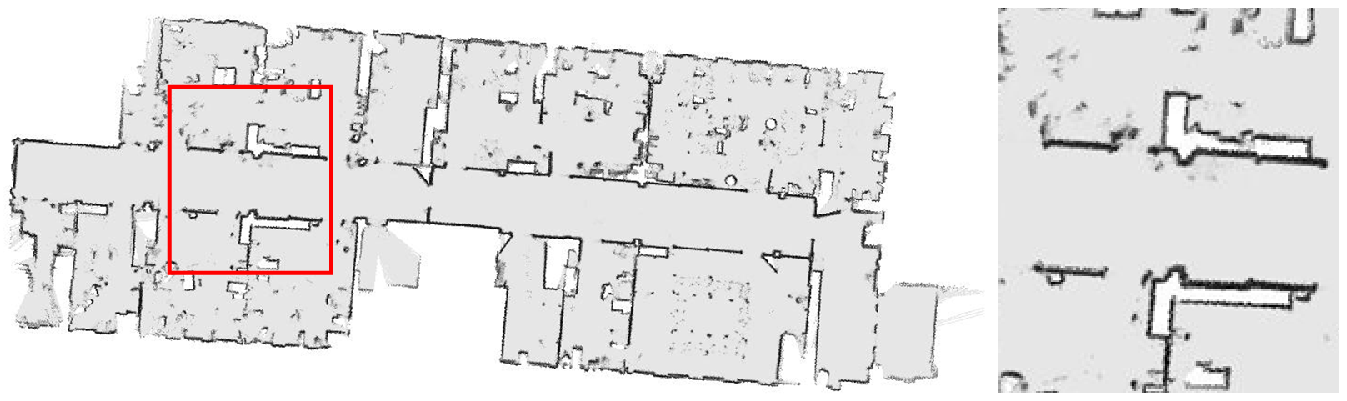}}
\end{center}
\caption{ The merging of multiple grid maps based different global motions. (a) Global motions estimated from the relative motions involved in the optimal MCS. (b) Global motions estimated from all reliable motions.}
\label{fig:ComFr}
\end{figure}

In one word, the proposed approach can accomplish the simultaneous merging of multiple grid maps with good accuracy.

\subsection{Comparison}

\begin{table}  
\renewcommand\arraystretch{1.5}         
\caption{Performance comparison for map merging of grid maps}
\centering                            
{\tabcolsep0.09in                     
\begin{tabular}{cccccccccc}
\toprule[1pt]
         & \multicolumn{3}{c}{Sequential merging \cite{Blanco13}} & & \multicolumn{3}{c}{Our method}\\
  \cline{2-4}\cline{6-8}
  Dataset & Obj. & T(s)           & Suc  & &  Obj.   & T(s) & Suc\\
  \hline
Tim &  1.4242  & 8.7890  &  Y  &   &  0.5546  &  5.0969   &  Y\\
Ineel & 16.2770 & 28.1093  &  Y  &   & 0.4509  &  20.4698   & Y\\
Fr079 & 4.7405  & 27.3639  &  N  &   & 0.2940   &  19.6659  & Y\\
\bottomrule [1pt]
\end{tabular}
}
\label{tab:com}
\end{table}

\begin{figure}[htp]
\begin{center}
\subfigure[]{\label{fig:ComFr-a}\includegraphics[scale=0.6]{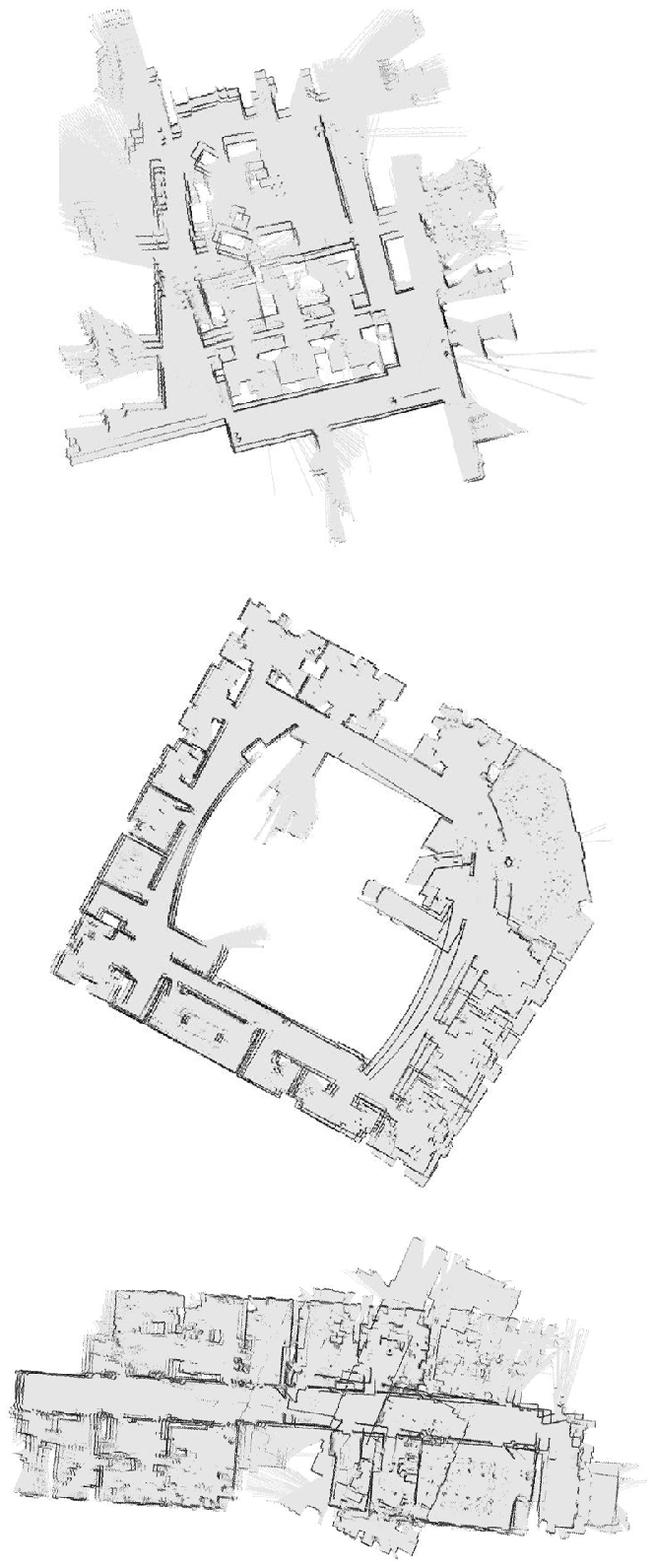}}
\subfigure[]{\label{fig:ComFr-b}\includegraphics[scale=0.6]{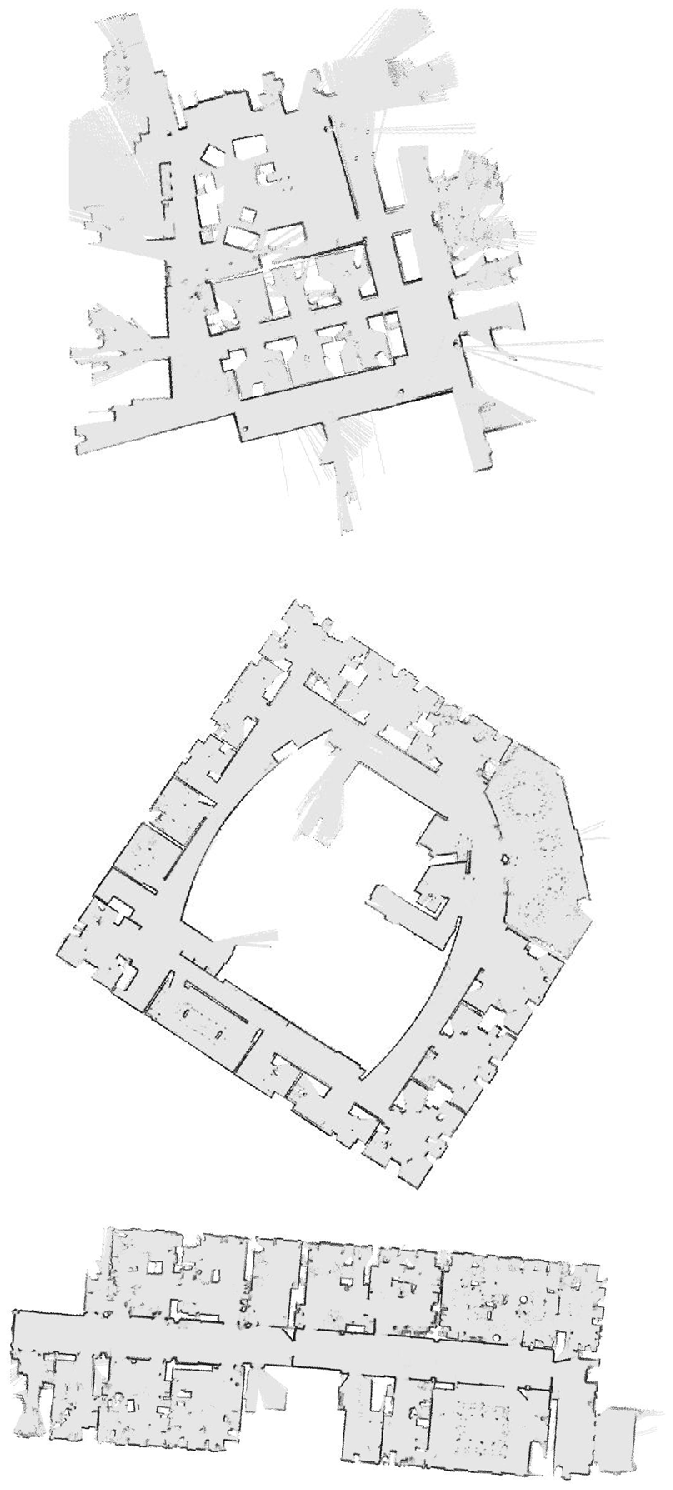}}
\end{center}
\caption{Multiple grid map merging results of three data sets for two competed approaches. (a) Results of sequential merging approach. (b) Results of the proposed approach.}
\label{fig:ComFr}
\end{figure}

To illustrate its superiority, the proposed approach requires to be compared with other related grid map merging approaches. However, to the best of our knowledge, few approaches can really accomplish the simultaneous merging of multiple grid maps. Therefore, the proposed approach is only compared with the sequential merging approach based on the pair-wise merging algorithm presented in \cite{Blanco13}. Experiments were tested on three grid map sets, which are displayed in Figs. \ref{fig:Tim}, \ref{fig:Fr079} and \ref{fig:Intel}, respectively. As there is no ground truth of global motions, the error criterion presented in \cite{Zhu16} can be utilized to quantitatively analyze the accuracy of competed merging approaches. During experiments, the runtime, merging error and merging status were recorded in Table \ref{tab:com}. To view the results in a more intuitive way, Fig. \ref{fig:ComFr} shows the merging results of three data sets for two competed approaches. As shown in Tabel \ref{tab:com} and Fig. \ref{fig:ComFr}, the proposed approach can obtain more efficient and accurate merging results than that of the sequential merging approach.

To merge multiple grid maps, the sequential merging approach estimate the relative motion of two grid maps and integrate them into one grid map, which will further be merged with another new grid map. The process of estimation and merging is repeated until all the grid maps are integrated into one global grid map.
Although this approach is straight-forward, it suffers from the well-known problem that merging errors accumulate at each step. As the grid map grows, the accumulate errors may lead to the failure of map merging. Therefore, the sequential merging approach can not always accomplish the merging of multiple grid maps. Besides, this approach requires to repeatedly extract SIFT features from the new merged grid map, so it is less efficient.

However, the proposed approach only utilizes the pair-wise merging approach to estimate relative motions of several map pairs. Among these estimated relative motions, there may exist unreliable ones. Subsequently, it randomly samples a minimum set of relative motions to estimate the initial global motions, which can be further confirmed by all relative motions. By repeating the process of sampling and confirming, it can find the optimal MCS for the estimation of initial global motions and confirm all reliable relative motions. Given the initial global motions, the motion averaging algorithm can be applied to all reliable relative motions so as to calculate the accurate global motions for simultaneous merging of multiple grid maps. Hence, the proposed approach can always accomplish merging multiple grid maps with good efficiency and accuracy.

\subsection{Robustness to grid map orders}
\begin{figure}[htp]
\begin{center}
\subfigure[]{\label{fig:ComFr-a}\includegraphics[scale=0.45]{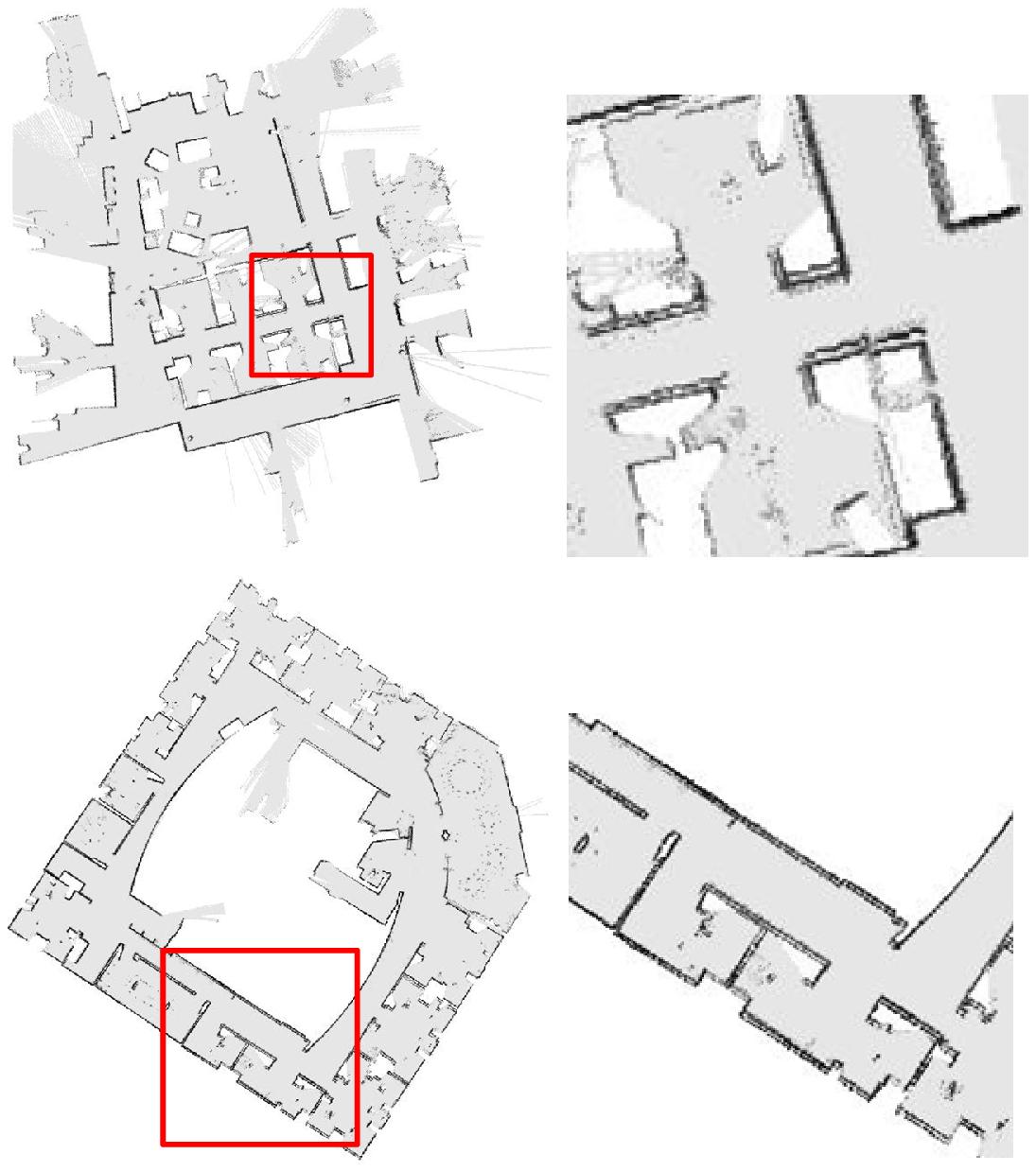}}
\subfigure[]{\label{fig:ComFr-b}\includegraphics[scale=0.45]{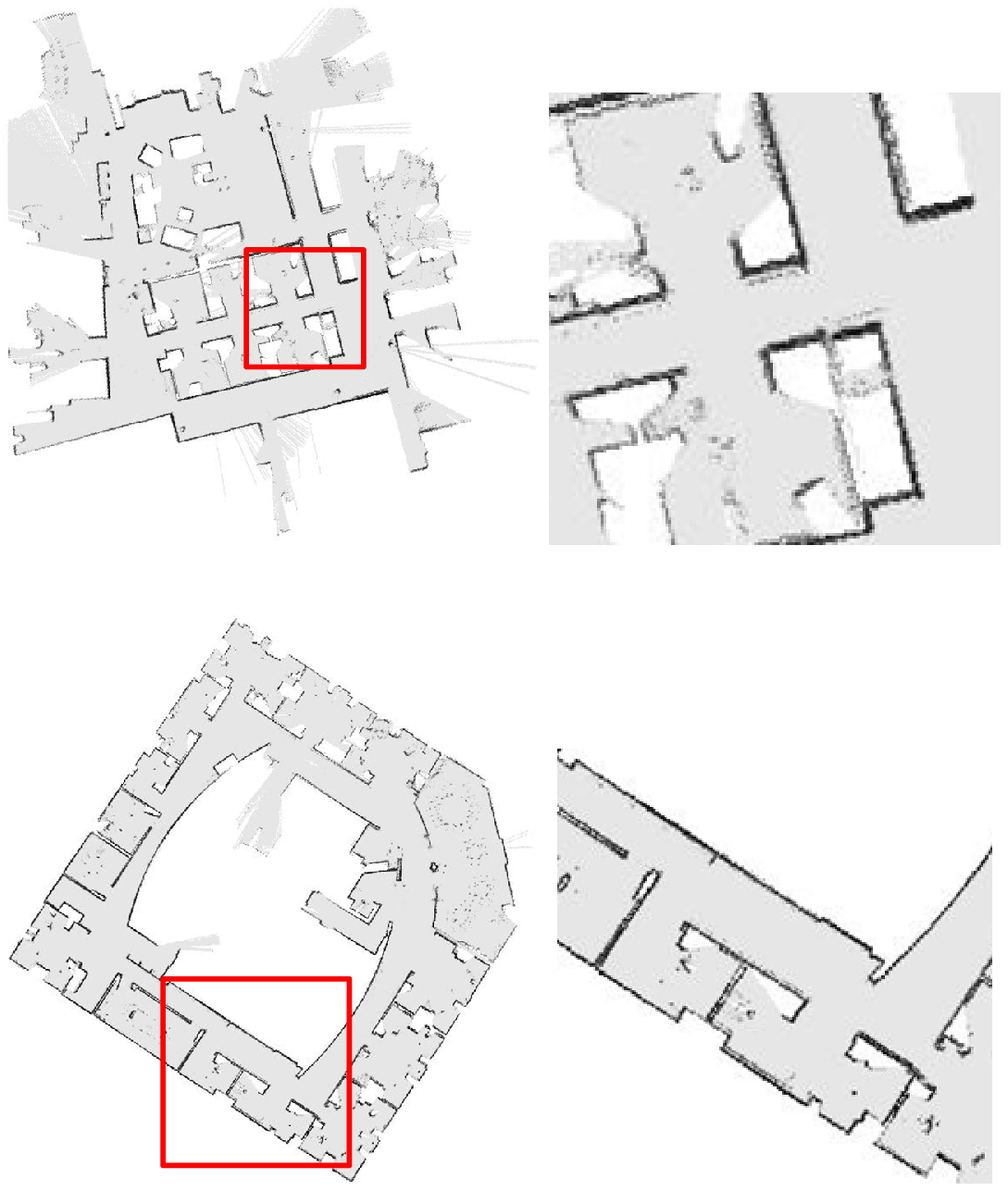}}
\end{center}
\caption{ Merged maps of Tim.log and Intel.log. (a) Merged maps based on initial global motions recovered from the optimal MCS. (b) Merged maps based on the refined global motions.}
\label{fig:Resu}
\end{figure}

To verify its robustness, the proposed approach was tested on three data sets with different group of orders, which can be randomly changed. During the experiment, grid maps with different orders were viewed as inputs and four groups of map merging results for each data set were recorded in Table \ref{tab:rob}. To view the results in a more intuitive way, Fig. \ref{fig:Resu} displays the merged maps for both Tim.log and Intel.log under one group of grid map order.
As shown in Table \ref{tab:rob}, the running time of the proposed approach is varied due to the size of grid map set. Besides, for each data set, the proposed approach can obtain almost the same merging results for different map orders.

\begin{table}  
\renewcommand\arraystretch{1.5}         
\caption{Map merging results for the grid maps with different orders.}
\centering                            
{\tabcolsep0.16in                     
\begin{tabular}{ccccccccccc}
\toprule[1pt]
    Dataset  &    ID  & \multicolumn{2}{c}{Error} & \multicolumn{2}{c}{T(s)} & Suc. \\
  \cline{3-6}
     &   &(Coarse) & (Fine) & (Coarse) & (Fine)  & \\
  \hline
Tim &  Order1  & 0.5713  &  0.5546  &4.7392	&0.3577  &Y \\
    &  Order2  & 0.5715  &  0.5560  &4.3052	&0.3587  &Y \\
    &  Order3  & 0.5713  &  0.5576  &4.5370	&0.3575  &Y \\
    &  Order4  & 0.5874  &  0.5497  &4.0893	&0.3582  &Y \\
    \hline
Intel & Order1  & 0.4877  &  0.4509  &19.1101 &1.3597 &Y \\
    &  Order2  & 0.4906  &  0.4470  &18.8025	&1.3739 &Y \\
    &  Order3  & 0.4882  &  0.4390  &19.663	&1.3715 &Y \\
    &  Order4  & 0.4944  &  0.4560  &18.6518	&1.3672  &Y \\
    \hline
Fr079 &  Order1  & 0.3083  &  0.2940  & 18.1678 & 1.4981 &Y \\
    &  Order2  & 0.3110  &  0.2938  &   18.4086 & 1.5012 &Y \\
    &  Order3  & 0.3084  &  0.2933  &  19.4578 & 1.5157  &Y \\
    &  Order4  & 0.3042  &  0.2936  &  17.6649 & 1.4938 &Y \\
\bottomrule [1pt]
\end{tabular}
}
\label{tab:rob}
\end{table}

 Before performing multiple gird merging, an exhaustive search strategy is utilized to independently estimate the relative motions of map pairs, and the results can be utilized to construct a undirected graph with all grid maps. On this constructed graph, a set of MCS are randomly sampled and then confirmed by all other relative motions. Subsequently, no matter what the order of grid maps is, the proposed approach can always search for the optimal MCS and obtain all the reliable relative motions. Based on the optimal MCS, it is easy to estimate good initial global motions. As shown in Fig \ref{fig:Resu}, initial global motions are not very satisfactory, so they can further be refined by the motion averaging algorithm with all the reliable relative motions. As shown in Table \ref{tab:rob}, the motion averaging only costs a small portion of merging time but can seriously reduce the merging error. Accordingly, the proposed approach can always obtain the grid map merging results, which are independent with the order of grid maps to be merged. Therefore, the proposed approach is robust to the order of grid maps to be merged.

\section{Conclusion}

This paper is, to the best of our knowledge, the first that proposes an effective approach for simultaneous merging grid maps built by multiple robots. Given a set of grid maps to be merged, it can accomplish grid map merging by several steps. It first utilizes the pair-wise map merging method to estimate the relative motion of grid map pairs. For the reason of low overlapping percentage, it may get unreliable estimation of relative motions for some grid map pairs. Therefore, the minimum set of reliable relative motions should be sampled and confirmed by other relative motions so as to eliminate unreliable relative motions. Then, the initial global motions can be estimated from the minimum set of reliable relative motions. Since the unreliable relative motions have been discarded, the motion averaging algorithm can be applied to the reserved relative motions so as to get accurate global motions for grid map merging. The proposed approach has been implemented and tested on the real robot data sets. Experimental results illustrate that the proposed approach can accomplish simultaneous merging multiple grid maps merging with good accuracy, efficiency and robustness.

The proposed approach includes some limitations. If one grid map has low overlap percentages with all other grid maps, it is difficult to obtain good pair-wise merging results for this grid map. In this case, there is no way to integrate it into the global grid maps. However, we note that most merging approaches proposed so far share this limitations as well. Besides, if these grid maps to be merged are in different resolutions, the proposed approach can not accomplish the merging of multiple grid maps. Our future work will focus on addressing the second limitation.

\section*{Acknowledgments}
This work is supported by the National Natural Science Foundation of
China under Grant nos. 61573273, 61573280 and 61503300.




\end{document}